\documentclass[conference]{IEEEtran}

\IEEEoverridecommandlockouts
\usepackage{cite}
\usepackage{amsmath,amssymb,amsfonts}
\usepackage{algorithmic}
\usepackage{graphicx}
\usepackage{textcomp}
\usepackage{xcolor}
\def\BibTeX{{\rm B\kern-.05em{\sc i\kern-.025em b}\kern-.08em
    T\kern-.1667em\lower.7ex\hbox{E}\kern-.125emX}}

\begin{document}

\title{Comparative Analysis of Image, Video, and Audio Classifiers for Automated News Video Segmentation\\
\thanks{This work is part of the project `Data-Informed Media Analysis Suite’ (DIMAS). It is financed by Xjenza Malta and The Malta Digital Innovation Authority for and on behalf of the Foundation for Science and Technology through the FUSION: R\&I Thematic Programmes, Digital Technologies Programme. This paper will be published in the IEEE 2025 Conference on Artificial Intelligence CAI2025 proceedings.}
}

\author{\IEEEauthorblockN{Jonathan Attard}
\IEEEauthorblockA{\textit{Department of Artificial Intelligence} \\
\textit{University of Malta}\\
Msida, Malta \\
jonathan.a.attard@um.edu.mt}
\and
\IEEEauthorblockN{Dylan Seychell}
\IEEEauthorblockA{\textit{Department of Artificial Intelligence} \\
\textit{University of Malta}\\
Msida, Malta \\
dylan.seychell@um.edu.mt}
}

\maketitle

\begin{abstract}

News videos require efficient content organisation and retrieval systems, but their unstructured nature poses significant challenges for automated processing. This paper presents a comprehensive comparative analysis of image, video, and audio classifiers for automated news video segmentation. This work presents the development and evaluation of multiple deep learning approaches, including ResNet, ViViT, AST, and multimodal architectures, to classify five distinct segment types: advertisements, stories, studio scenes, transitions, and visualisations. Using a custom-annotated dataset of 41 news videos comprising 1,832 scene clips, our experiments demonstrate that image-based classifiers achieve superior performance (84.34\% accuracy) compared to more complex temporal models. Notably, the ResNet architecture outperformed state-of-the-art video classifiers while requiring significantly fewer computational resources. Binary classification models achieved high accuracy for transitions (94.23\%) and advertisements (92.74\%). These findings advance the understanding of effective architectures for news video segmentation and provide practical insights for implementing automated content organisation systems in media applications. These include media archiving, personalised content delivery, and intelligent video search.

\end{abstract}


\section{Introduction}\label{introduction}

News videos are becoming more relevant as video technologies are becoming more widespread, especially in news content \cite{zhu2001automatic} \cite{seychell24}.  News videos offer a unique combination of rich visual and auditory information, making them a powerful tool for conveying complex narratives and engaging audiences globally. However, the rapid expansion in the volume of video content presents considerable challenges in terms of organisation, retrieval, and analysis, mainly when dealing with large and unstructured datasets.

Although news videos are a convenient medium, navigating their content remains time-intensive. Automated tools have the potential to transform this experience by enabling users to interpret content efficiently and reducing the time required to search for relevant information \cite{gao2002unsupervised}. Research focuses on methods for summarising, organising, and categorising video content using computer vision and related technologies. A structured and condensed news video database is crucial, as it is impractical for individuals to watch all programmes across various channels indiscriminately \cite{gao2002unsupervised}. 

The issue of information overload further highlights the challenges associated with video content consumption. Video data's vast and unstructured nature can hinder consumers’ ability to extract meaningful insights, diminishing their capacity to make informed decisions. As discussed by Edmunds and Morris, this could negatively impact effective decision-making due to the lack of informed decisions \cite{edmunds2000problem}.

One of the most common initial steps for video content organisation is segmenting the video into different meaningful parts \cite{hauptmann1998story}. Proper segmentation facilitates content organisation, improves accessibility, and enhances the ability to perform analytics on large video datasets \cite{hauptmann1998story}. It also enables users to navigate content more efficiently, extracting specific information from lengthy broadcasts without viewing them in their entirety. Despite its importance, traditional approaches to segmentation, which often involve manual effort, are highly labour-intensive and unsuited for managing the scale of modern video datasets \cite{zhu2001automatic}. Recent advancements in computer vision and machine learning have advanced the development of automated segmentation techniques. These methods leverage multimodal data, such as visual cues, audio signals, and textual elements like captions, to identify boundaries between distinct segments in a video. Automated segmentation accelerates the processing of large volumes of video content and ensures consistency and accuracy in the segmentation process.

The primary aim of this study is to establish a framework for automating video segmentation and scene classification, particularly in news videos. The initial objective involves collecting and annotating a diverse set of news videos. The research focuses on training and evaluating models ranging from basic image classifiers to advanced multi-modal classifiers, including visual, temporal, and audio data. This study seeks to provide a comprehensive understanding of the most effective scene segmentation methods, from the annotation and training processes to the outcomes of accurately segmented news videos.  

This paper begins with a background and literature review in section \ref{background}, examining existing technologies and studies related to news video segmentation and classification, digital video libraries, news video analysis, and content retrieval systems. Section \ref{methodology} outlines the method to achieve the aims and objectives, including the implementation details and rationale behind each step. The results are presented in section \ref{results}, discussing the outcomes of the proposed methodology, with a detailed comparison of models, highlighting the benefits and limitations. Finally, section \ref{conclusion} concludes with a summary of the study, emphasising key findings and suggesting potential avenues for future research and improvements.

\section{Related Work}\label{background}

This section provides an overview of related studies on video segmentation, focusing on their applications to news videos. A detailed evaluation of video classification techniques is included, as various classifiers will be tested and compared in this study to identify the most reliable methods. Additionally, the challenges and limitations associated with video segmentation and classification are discussed, including brief descriptions of the approaches used by previous studies to achieve segmentation and classification for videos to establish the context for this research.

Scene segmentation is a method of dividing a video into different labelled parts. This method could have multiple applications, especially in content organisation. A standard approach to achieve this is to analyse individual frames and subsequently merge consecutive frames with similar labels to form coherent scenes. For instance, Rafiq et al. \cite{rafiq2020scene} introduced a frame-based approach for sports video classification, reporting a remarkable accuracy of 99.26\%. Their methodology extracted six frames per second, each resized to dimensions of 227x227×3 and utilised an AlexNet CNN \cite{krizhevsky2012imagenet}. Using the pre-trained weights of ImageNet \cite{deng2009imagenet}, transfer learning was applied to adapt the model for five specific classes.

Domain-specific knowledge has proven to be a reliable approach for news video segmentation, as highlighted by Gao and Tang \cite{gao2002unsupervised}. For example, Hauptmann and Witbrock \cite{hauptmann1998story} used patterns such as audio silence, acoustic noise, and black frames to identify segment boundaries. Similarly, Gong et al. \cite{gong1995automatic} classified football videos by analysing the field layout, demonstrating the effectiveness of structural cues. Zhu and Liou \cite{zhu2001automatic} emphasised the role of speech over visual information, identifying keywords to segment and classify news videos into eight categories, including Science and Technology, Business, Health, Entertainment, Weather, Sports, Daily Events, and Politics. However, this approach assumes the availability of reliable pre-trained transcription tools, which are not universally accessible, particularly when considering uncommon languages such as Maltese. Hauptmann and Witbrock addressed this by using closed captions when available or employing speech recognition to generate them, particularly for tasks like advertisement detection. 

An alternative approach involves video recognition techniques that leverage temporal data between frames, yielding effective results. For instance, MovieCLIP \cite{bose2023movieclip} integrates scene boundary detection with video recognition. Specifically, PySceneDetect \cite{pyscenedetect}, a Python library, is employed to partition the video into discrete segments. Each segment is subsequently analysed on a shot-by-shot basis using the CLIP model, which is particularly adept at zero-shot labelling tasks. As video segmentation and classification are often treated as separate processes, the subsequent subsection will focus on studies dedicated to video classification.

\subsection{Video Classifiers}

Video classification has advanced significantly over time, transitioning from early frame-based methods utilising CNN architectures like AlexNet \cite{krizhevsky2012imagenet}, VGGNet \cite{simonyan2014very}, and ResNet \cite{he2016deep} to more sophisticated techniques such as 3D CNN \cite{carreira2017quo}, RNN \cite{donahue2015long}, and ViT \cite{arnab2021vivit}. These innovations have enhanced the capacity to capture spatio-temporal features, improving classification performance. Multi-modal approaches have emerged as an alternative to purely visual methods, integrating audio \cite{li2022mvitv2} and text-based features \cite{wu2023newsnet}. However, challenges such as missing transcripts in specific contexts and the linguistic variability of text data remain unresolved, particularly in multilingual scenarios.

Notable progress for video classification in a temporal fashion includes the SlowFast network \cite{feichtenhofer2019slowfast}, which employs dual pathways to separately process spatial and temporal information, demonstrating state-of-the-art performance. The X3D model \cite{feichtenhofer2020x3d} builds upon this foundation, achieving comparable accuracy with reduced computational demands. Transformers have also significantly influenced video classification. Vision-based transformers, such as ViViT \cite{arnab2021vivit}, extend the capabilities of ViT by modelling temporal dependencies, surpassing prior 3D CNN-based methods. Regularisation strategies and uniform frame sampling have further contributed to the effectiveness of such models. Modern approaches, like MViTv2 \cite{li2022mvitv2}, employ attention pooling to handle multiple frames simultaneously, achieving high accuracy on datasets like Kinetics-400. Nevertheless, these models demand substantial computational resources, with requirements such as 8 V100 GPUs. Although this, surprisingly a study by Kareer et al. \cite{kareer2024we} revealed that their application of image classifiers strongly outperformed the video classifiers despite having the temporal element as an advantage.

\begin{figure}[htbp]
  \centering
  \includegraphics[width=0.4\textwidth]{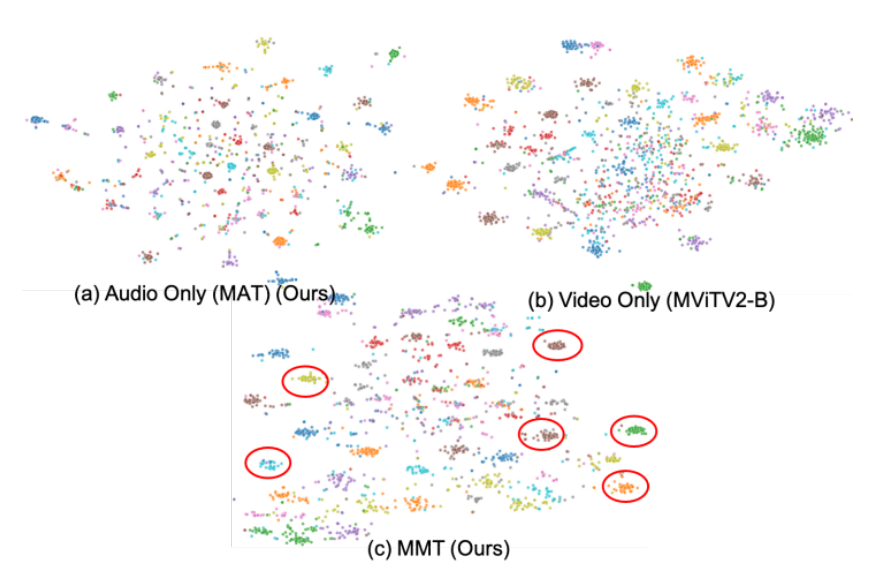}
    \caption{Visualisation of audio-visual representation, demonstrating how multimodal models can collaboratively capture additional features to represent features better (Source: \cite{wu2023newsnet})}
  \label{fig:audio_visual_transformers}
\end{figure}

Multi-modal techniques have been explored, combining visual, audio, and textual inputs for richer video representation \cite{zhu2024efficient}. For instance, Zhu et al. \cite{zhu2024efficient} employed a multi-model approach integrating an audio spectrogram transformer, visual transformer, and bottleneck mechanism, achieving superior classification results since it can better represent the video data as shown in Figure \ref{fig:audio_visual_transformers}, despite significant resource requirements.

Since video classification can be done using both video classifiers and audio classifiers combined, as shown by different studies \cite{zhu2024efficient}, it is essential to explore other audio classifiers as well. Audio classification involves analysing audio signals and categorising them into predefined labels. As an integral component of video data, audio classification enables a better understanding of video content through applications such as speech, music, environmental sound, and natural language processing \cite{kaur2021audio}. Early methods employed machine learning techniques like SVMs, KNN, and ANN \cite{zaman2023survey}. However, deep learning models, particularly those leveraging automatic feature extraction, have since become the dominant approach \cite{cui2022research}. Advances include the adoption of CNN, RNN, transformers, and hybrid models, which have collectively improved classification performance \cite{zaman2023survey}.

Audio data is often transformed into alternative representations such as spectrograms, MFCC, or waveforms to enhance classification accuracy. These representations, including mel spectrograms, chromograms, and tempograms, facilitate the application of image-based classifiers \cite{khan2020survey}. Recent work by Lui et al. introduced DiffRes, a mel spectrogram compressor, which improves classification accuracy by optimising the information density of spectrograms \cite{liu2024learning}.

State-of-the-art methods include the Audio Spectrogram Transformer (AST), which adapts the ViT architecture to process mel spectrograms more efficiently \cite{gong2021ast}. The AST model improves classification accuracy and computational efficiency by reducing input dimensionality and leveraging transfer learning. Subsequent advancements, such as the Multi-Scale Audio Spectrogram Transformer (MAST) \cite{zhu2023multiscale}, introduced parameter reductions via pooling layers and cross-embedding attention mechanisms, resulting in further accuracy improvements while maintaining the lightweight architecture. These developments highlight the growing potential of transformers in advancing audio classification.

\section{Methodology}\label{methodology}

The methodology involves collecting and annotating news video data and training different classifiers to predict scene types. These classifiers are compared, and an automated system is developed to segment and label the videos, providing a user interface for browsing. Key analytics, such as scene frequency, are derived alongside classification performance.

\subsection{Dataset}
\begin{figure}[htbp]
    \centerline{\includegraphics[width=0.5\textwidth]{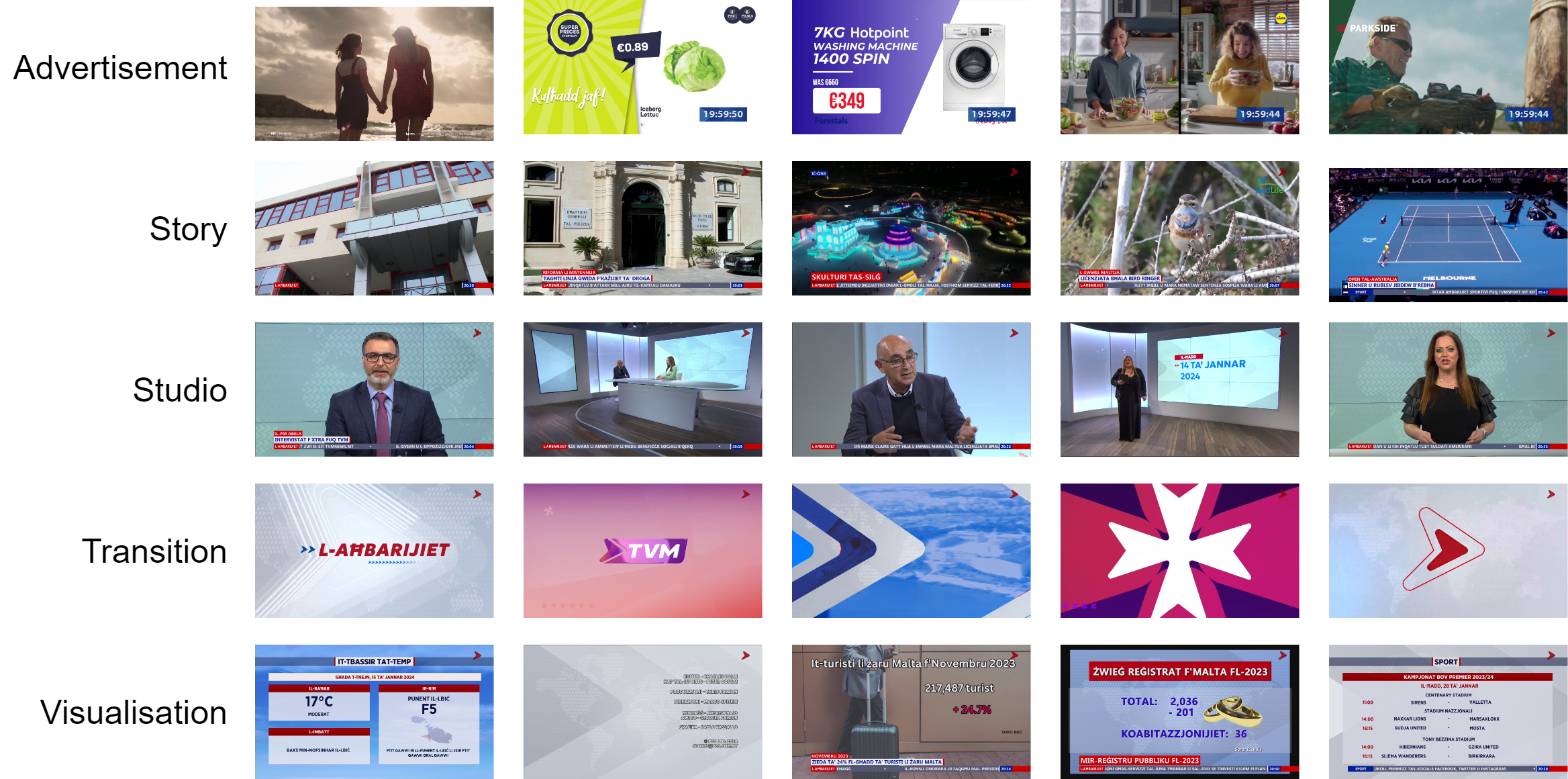}}
    \caption{Example frame samples corresponding to each of the five scene classification labels: Advertisement, Story, Studio, Transition, and Visualisation.}
    \label{fig:scene_examples}
\end{figure}

The dataset includes 41 news videos, averaging 50 minutes each, manually annotated by third-party independent annotators. The videos were classified into five scene types: Advertisements, Stories, Studio scenes, Transitions, and Visualisations, as shown in Fig. \ref{fig:scene_examples}.

Selecting an annotation tool for such a task is an essential step, as providing annotators with an easy-to-use tool will help to ensure accurate annotations. However, few tools tackle video segmentation and scene labelling as a single task. Label Studio \cite{labelstudio}, an open-source tool, was selected for video annotation due to its customisable interface. The interface was modified to include a video preview with a timeline, allowing annotators to move frame-by-frame and accurately label scenes. To reduce server load, the video resolution was set to 384x216px, also used for model training. A total of 1,832 scene clips were annotated from 41 videos and split into training, validation, and testing sets.

\begin{figure}[htbp]
    \centerline{\includegraphics[width=0.5\textwidth]{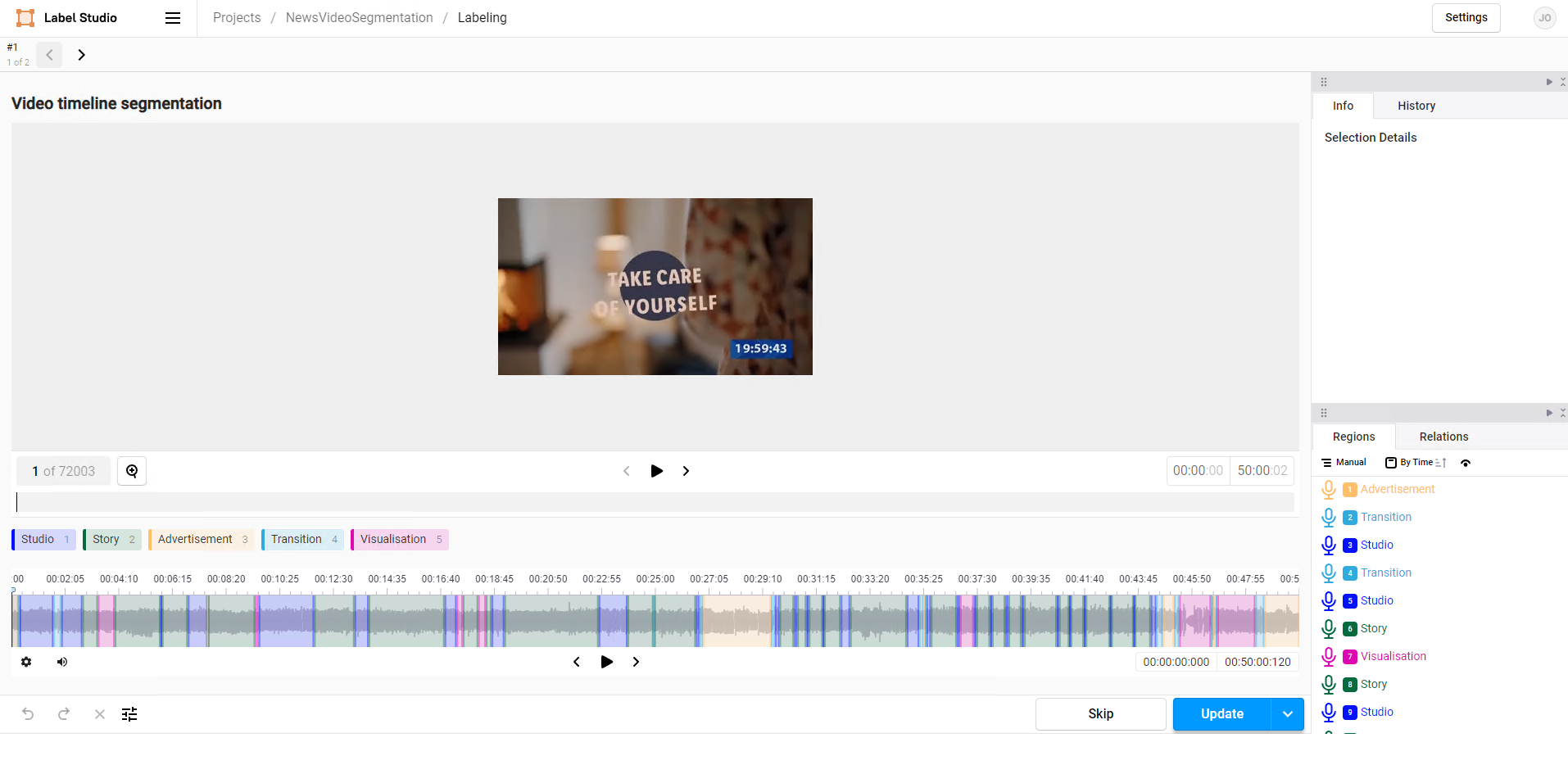}}
    \caption{Interface used for video annotation, demonstrating a video fully annotated. Users had the option to move frame-by-frame for accuracy, jump, highlight, and label different sections while viewing and listening to the video}
    \label{fig:annotation}
\end{figure}

\subsection{Scene Classification}

For video classification, four models were trained to evaluate performance. These included image classifiers, multi-frame video classifiers, audio classifiers, and multi-modal classifiers. Python and PyTorch are used for dataset creation and model training. Due to the lack of GPU support for TensorFlow on Windows, PyTorch is preferred. The study will be conducted on a machine with 32 GB of RAM and a GeForce RTX 4090, suitable for training larger models, though not as extensive as those in other studies  \cite{zhu2024efficient, li2022mvitv2}. 

Video classification was approached by extracting and classifying individual frames, as demonstrated by Rafiq et al. \cite{rafiq2020scene} by using AlexNet \cite{krizhevsky2012imagenet}. In addition to AlexNet, ResNet152 \cite{he2016deep} will be used as baseline architectures. Sixteen frames were uniformly sampled per scene, with an additional 10-frame padding at both ends to minimise annotation errors near scene transitions, requiring a minimum of 36 frames per clip. A dedicated dataset of 28,817 images was created to enhance computational efficiency during training. Augmentation techniques, including resizing images to 50-100\% of their original size, maintaining aspect ratios between 0.75 and 1.33, and applying random horizontal flips with a 50\% probability, enhanced data diversity. Weighted sampling addressed class imbalances. Both models utilised pre-trained ImageNet weights. Input images resized to 224×224px, the Adam optimiser with a learning rate of 0.0001, and cross-entropy loss. Training was conducted with a batch size of 32 for up to 100 epochs, incorporating early stopping with a patience of 15 epochs. ResNet152’s deeper architecture, featuring 152 layers and residual learning, was evaluated alongside AlexNet to benchmark performance against advanced models \cite{ullah2022comparative, nyarko2022comparative}.

Video classification utilised a temporal-aware approach based on the ViViT model by Arnab et al. \cite{arnab2021vivit}, modified to reduce computational demands. The model's parameters were halved, reducing transformer layers (L=12 to L=6), attention heads (NH=12 to NH=6), and hidden dimensions (d=768 to d=384). Input frames were resized to 224×224 pixels, and sequences of 16 frames, with a 10-frame padding, were selected for processing. Pre-trained weights from the Kinetics 400 dataset were employed, though the parameter reduction limited their efficacy. Augmentation techniques, including random resizing, cropping, and horizontal flipping, were done with the image classification; however, in this case, the same augmentation was applied to all images within each clip. Training used a batch size of 16 with 16 frames per clip, the AdamW optimiser with a learning rate of 0.005, and 23 workers to optimise throughput. To train the models, 16 frame sequences were extracted randomly from annotated videos, ensuring at least 10-frame padding from the start and end of videos. The training was configured for up to 2000 epochs but was manually terminated once improvements plateaued. 

Audio classification employed the AST model, introduced by Gong et al. \cite{gong2021ast}, demonstrating state-of-the-art performance in audio-based tasks. To ensure consistency with the ViViT model for video classification, the AST parameters were halved, reducing the transformer layers from 12 to 6 (L=6), attention heads from 12 to 6 (NH=6), and hidden dimensions from 768 to 384 (d=384). Mel spectrograms were chosen for input representation due to their effectiveness in capturing key audio features, as highlighted in Gong et al. \cite{gong2021ast}. Spectrograms were generated with a sampling rate of 44,100 Hz, a window size 2048, a hop length 512, and 128 mel bins, ensuring high-quality audio feature representation. To align audio and video durations, each audio clip corresponded to 16 video frames, requiring 27,648 audio frames per clip, with this number calculated to maintain temporal consistency between modalities. Pre-trained AST weights fine-tuned on AudioSet were employed to leverage prior knowledge \cite{gong2021ast}, although the parameter reduction limited their full utility. The training was conducted with a batch size of 16, where each batch processed one 16-frame audio clip using the AdamW optimiser with the same parameters of ViViT training. The training process was set for 2000 epochs and manually terminated upon performance stabilisation, ensuring efficient and temporally consistent processing.

\begin{figure}[htbp]
    \centerline{\includegraphics[width=0.5\textwidth]{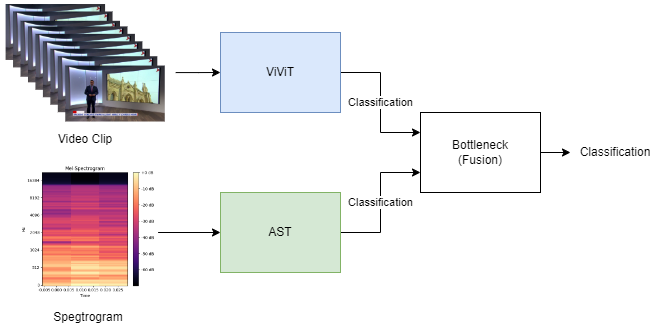}}
    \caption{Multi-modal architecture showing the interaction of the vision and audio models combined through a fusion layer.}
    \label{fig:multi_modal}
\end{figure}

The multi-modal classification architecture developed for this study integrates both video and audio models, inspired by Zhu’s work \cite{zhu2024efficient} on combining the MViTv2 and MAT models. This approach leverages the base ViViT and AST models instead of the more complex variants to balance simplicity, resource efficiency, and accuracy. The outputs of the video and audio models are fused via a concatenation method as shown in Fig. \ref{fig:multi_modal}, simplifying the process and enhancing computational efficiency compared to the AV bottleneck used in Zhu’s framework \cite{zhu2024efficient}. For video, the ViViT model processes video frames, and for audio, the AST model classifies mel spectrograms, while the fusion layer consolidates the outputs into a final classification. Synchronisation between video frames and audio is crucial to avoid misalignment, which could lead to inaccurate model outputs. The models are trained from scratch, not using previously trained weights. Two variations of the multi-modal model are tested: ViViT-AST, using 16 frames, and ViViT-AST-L, using 32 frames to provide a larger temporal context.
Additionally, individual binary models will be trained on the labels from the dataset using a one-vs-all approach for more focused learning. The model training for both configurations uses transfer learning, employing pre-trained weights from Kinetics 400 and AudioSet for the video and audio components, respectively. The batch size for the ViViT-AST model is set at 16, while for the larger ViViT-AST-L model, the batch size is reduced to 8 due to the higher computational load. 

\subsection{Generating Results}  

Model training involved calculating test and validation accuracies based on clip-level class predictions. However, for evaluation, a two-step approach was employed to segment raw videos into labelled parts automatically. First, scene detection was performed using PySceneDetect \cite{pyscenedetect}, a widely adopted Python library \cite{bose2023movieclip}, to divide the video into unlabelled scenes. These scenes were then classified using the trained models, while neighbouring scenes with the same predicted class were merged into a single segment.  

The results were generated by comparing each model's annotated and predicted scene timelines, producing a multi-class confusion matrix. This matrix recorded the durations of correct and incorrect classifications, facilitating the computation of precision, recall, and accuracy metrics. The confusion matrices can also reveal which classes are more reliable than others across the different trained models. This two-step approach demonstrates a reliable method, supported by evaluation results, on how videos can be automatically segmented into different labelled scenes.

\section{Results}\label{results}

This section discusses the results produced by the trained models, including any challenges and limitations of the current methodology. First, an overview of the annotated dataset will be discussed, specifically mentioning the class distribution within the trained dataset.

The label distributions of the annotated dataset are presented in Table \ref{tab:label_distribution}. Story scenes had the longest average duration, followed by advertisements, while transitions were the shortest. Advertisements and visualisations were the least frequent, whereas stories and studio scenes were the most numerous. These scene count and duration imbalances impacted model training and presented challenges in achieving adequate sample balancing.

\begin{table}[htbp]
\caption{Class distribution of the scene clips used for training, including the number of clips, average duration, and total duration for each label.}
\begin{center}
    \begin{tabular}{|l|c|c|c|}
    \hline
    \textbf{Label} & \textbf{Clip No.} & \textbf{Avg. Dur. (Secs)} & \textbf{Total Dur. (Hrs)} \\ \hline
    Advertisement & 126 & 62.75 & 2.20 \\ \hline
    Story & 631 & 78.03 & 13.68 \\ \hline
    Studio & 655 & 30.38 & 5.53 \\ \hline
    Transition & 295 & 7.82 & 0.64 \\ \hline
    Visualisation & 125 & 36.28 & 1.26 \\ \hline
    \end{tabular}
    \label{tab:label_distribution}
\end{center}
\end{table}

Table \ref{tab:model_performance} shows the model performance for each model tested. Note that these results include the scene detection stage and are calculated using the confusion matrices as bases described in the methodology. The most surprising outcome at first glance is that the image classifiers outperformed the more advanced models. Namely, the ResNet model achieved the best results. However, the ViViT achieved relatively close results, although given the memory and computation complexity, using the image classifiers would be far better. These results closely resemble the ones shown within the study by Kareer et al. \cite{kareer2024we}, where the image classifiers outperformed the video classifiers using a temporal element significantly.

\begin{table}[htbp]
\caption{Comparison of scene classification model performance, evaluated based on test accuracy, speed, training duration, and model efficiency. The bold values indicate the best-performing model for each metric.}
\begin{center}
    \begin{tabular}{|c|c|c|c|c|}
    \hline
    
    \textbf{Model} & \textbf{Acc. (\%)} & \textbf{Min. / Epoch} & \textbf{Train Dur. (Hrs)} \\ \hline
    AlexNet             & 84.00  &\textbf{0.45}                 & \textbf{0.08}             \\ \hline
    ResNet              & \textbf{84.34}  & 2.13        & 0.89                      \\ \hline
    ViViT               & 75.51   & 1.93                & 14.64                     \\ \hline
    AST                 & 52.93    & 4.13               & 14.73                     \\ \hline
    ViViT-AST         & 72.70       & 7.8             & 85.57                     \\ \hline
    ViViT-AST-L       & 67.48       & 28.76            & 90.59         
    \\ 
    
    \hline
    \end{tabular}
    \label{tab:model_performance}
\end{center}
\end{table}

A significant challenge in training the models was the memory and computational demands of the larger architectures. Even a minor error or adjustment during the training process often required restarting the training to ensure fairness in evaluation. Since each epoch required substantial time, some training processes extended over weeks. Notably, the ViViT-AST-L model lasted nearly four days straight to complete its training, significantly limiting the flexibility for testing and experimentation.

\begin{figure}[htbp]
    \centerline{\includegraphics[width=0.5\textwidth]{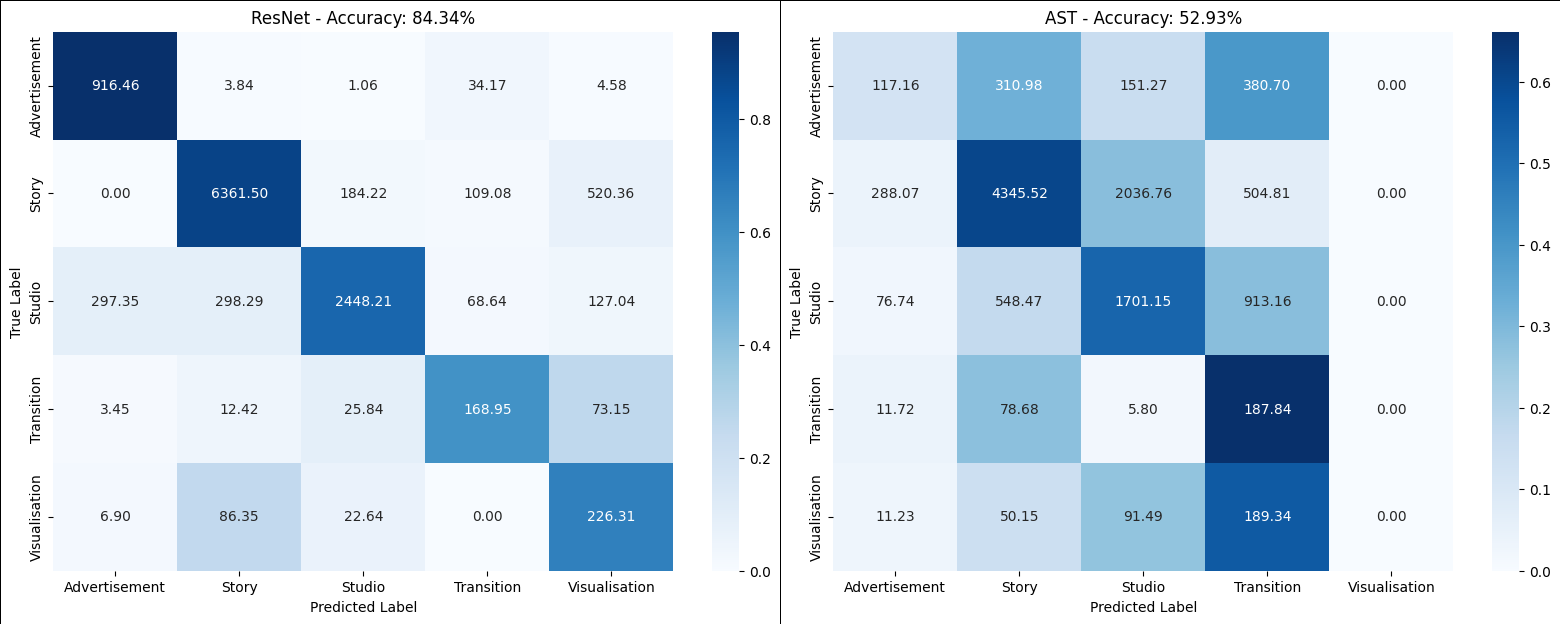}}
    \caption{Sample confusion matrices of the results generated from the ResNet and the AST models. Note that the other confusion matrices not presented closely resemble the patterns of the ResNet model.}
    \label{fig:confusion_matrices}
\end{figure}

The confusion matrices for each model reveal performance disparities in classifying scene types in news videos. AlexNet and ResNet excel in classifying "Story" and "Advertisement" scenes with minimal errors. The AST model struggles with "Advertisement" and "Visualisation" scenes, likely because visualisation scenes interrupt visuals while audio remains unchanged, making them challenging to detect. For "Advertisement" scenes, their audio similarity to other scenes might explain the difficulty. However, the AST model excels at recognising transitions due to their distinctive recurring sounds. Fig. \ref{fig:confusion_matrices} displays the confusion matrices for the general model combining ResNet and AST.

Binary classification models were fine-tuned using transfer learning from the pre-trained ViViT-AST-L model to distinguish the results for each class, with training times averaging 35 hours per model and exceeding one week for all five models. Performance results, summarised in Table \ref{tab:binary_results}, reveal that the Transition model achieved the highest accuracy, likely due to the recurring nature of transition scenes across the news videos. For "Studio" and "Visualisation," this aligns with the confusion matrix results, where the ViViT-AST-L model demonstrated lower reliability for these classes. However, while the "Story" model was less reliable than "Advertisement" and "Transition," it still showed sufficient accuracy within the confusion matrix.

\begin{table}[htbp]
\caption{Performance results of the binary classification models fine-tuned from the ViViT-AST-L model. The models distinguish each class Advertisement, Story, Studio, Transition, and Visualisation from all other classes in a one-vs-all fashion.}
\begin{center}
    \begin{tabular}{|l|l|l|l|l|l|}
    \hline
    \textbf{Bin. Cls} & Adv. & Story & Studio & Tran.     & Vis. \\ \hline
    \textbf{Acc. (\%)}         & 92.74                  & 80.12          & 78.97  & 94.23 & 82.15         \\ \hline
    \end{tabular}
    \label{tab:binary_results}
\end{center}
\end{table}

\section{Conclusion}\label{conclusion}

This study presented an AI-driven approach to news video segmentation in an automated manner, addressing the growing need for efficient tools to process and analyse large volumes of news video content. By combining scene detection and classification, this research demonstrates the results of several scene classification methods to achieve video segmentation automatically. This study evaluates the differences between classical image classifiers in scene classification against more robust models such as ViT, AST, and multi-modal approaches that integrate temporal, visual, and audio elements. Interestingly, despite their simplicity, image classifiers outperformed the more complex models in accuracy and computational efficiency. This finding suggests that further refinement is needed for the bigger models to take full advantage of their complexity. Additionally, the study underscores the lack of specialised annotation tools and the high resource demands of sophisticated models, which could present significant obstacles for future research with limited computational resources.

The results contribute to advancing video analysis methods, particularly in the media industry, where such tools can improve content navigation and support more comprehensive monitoring of news coverage. Future work will focus on refining segmentation techniques, applying more attention-based techniques \cite{seychell18}, expanding the dataset, and exploring other multi-modal approaches to enhance the robustness of scene classification. With continued development, these tools can significantly improve the efficiency and depth of video analysis in many fields, such as digital archives and news video consumption.

\bibliography{references.bib}

\end{document}